\documentclass[10pt,twocolumn,letterpaper]{article}
\usepackage[pagebackref=true,breaklinks=true,letterpaper=true,colorlinks,bookmarks=false]{hyperref}
\usepackage{cvpr}
\usepackage{times}
\usepackage{epsfig}
\usepackage{graphicx}
\usepackage{amsmath}
\usepackage{amssymb}

\pdfoutput=1
\cvprfinalcopy 

\def\cvprPaperID{****} 

\begin{document}

\title{Attentive Spatio-Temporal Representation Learning for Diving Classification}

\author{Gagan Kanojia \quad\quad Sudhakar Kumawat \quad\quad Shanmuganathan Raman\\
Indian Institute of Technology Gandhinagar, India\\
{\small\{\tt gagan.kanojia, sudhakar.kumawat, shanmuga\}@iitgn.ac.in}}

\maketitle

\begin{abstract}
Competitive diving is a well recognized aquatic sport in which a person dives from  a platform or a springboard into the water. Based on the acrobatics performed during the dive, diving is classified into a finite set of action classes which are standardized by FINA. In this work, we propose an attention guided LSTM-based neural network architecture for the task of diving classification. The network takes the frames of a diving video as input and determines its class. We evaluate the performance of the proposed model on a recently introduced competitive diving dataset, Diving48. It contains over 18000 video clips which covers 48 classes of diving. The proposed model outperforms the classification accuracy of the state-of-the-art models in both 2D and 3D frameworks by 11.54\% and 4.24\%, respectively. We show that the network is able to localize the diver in the video frames during the dive without being trained with such a supervision.
\end{abstract}
\section{Introduction}
\begin{figure*}[h!]
    \centering
    \includegraphics[width=0.12\linewidth]{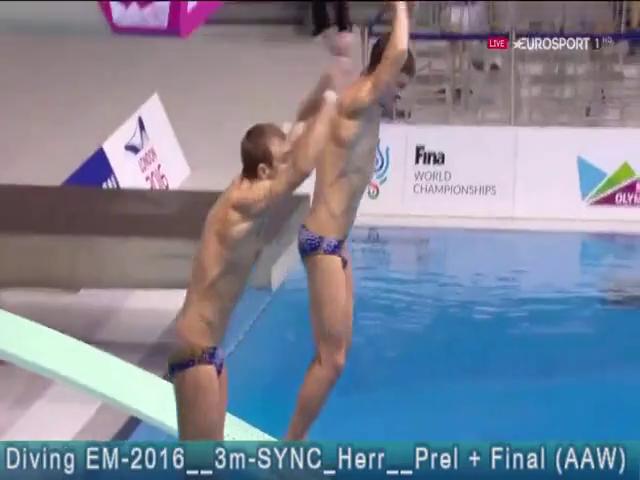}
    \includegraphics[width=0.12\linewidth]{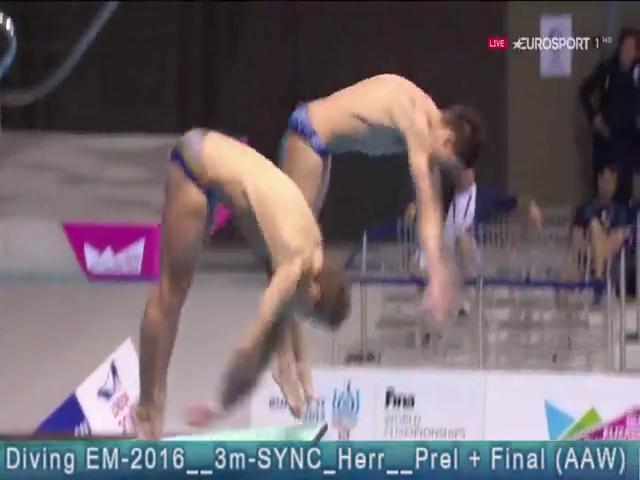}
    \includegraphics[width=0.12\linewidth]{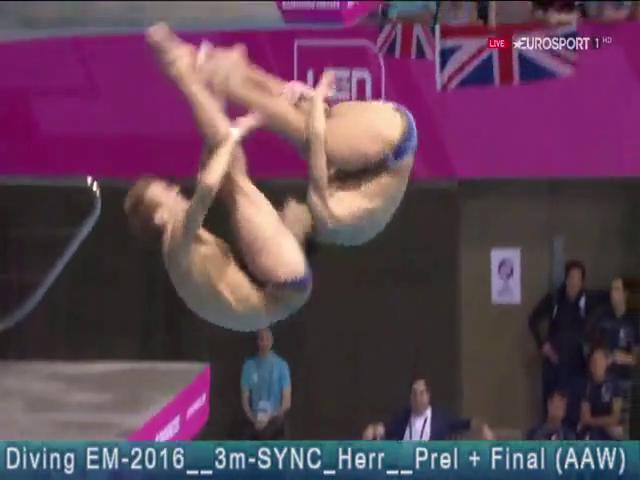}
    \includegraphics[width=0.12\linewidth]{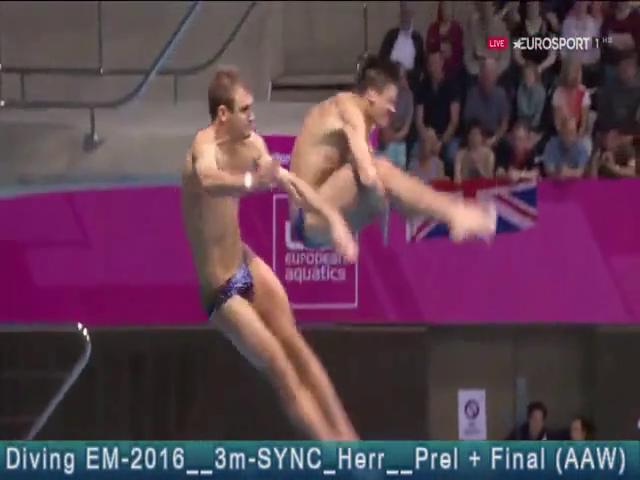}
    \includegraphics[width=0.12\linewidth]{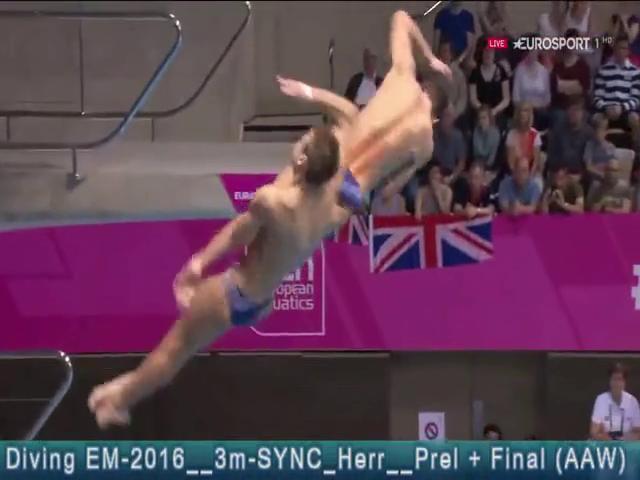}
    \includegraphics[width=0.12\linewidth]{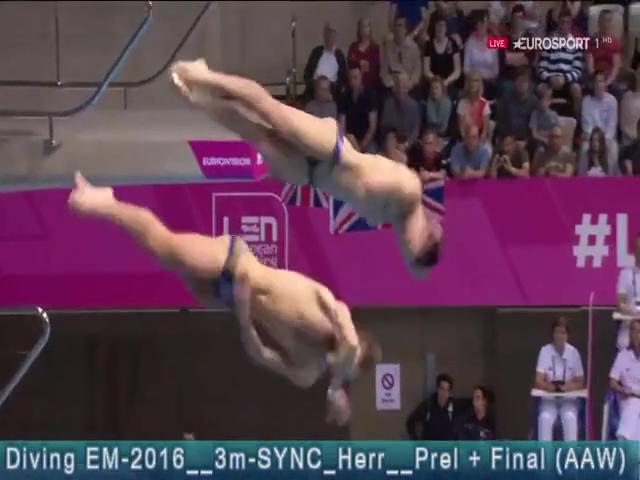}
    \includegraphics[width=0.12\linewidth]{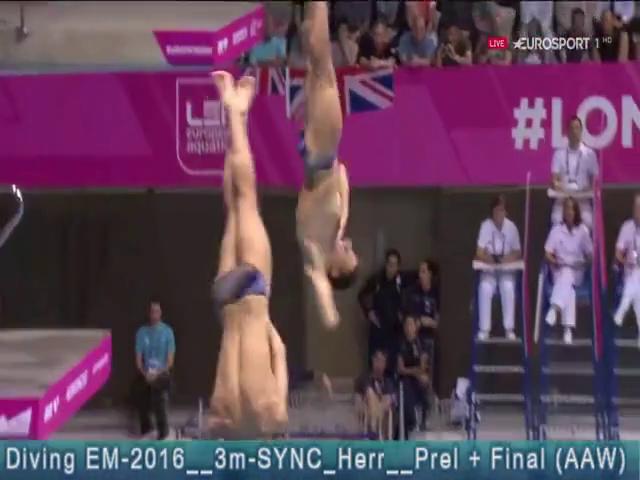}
    \includegraphics[width=0.12\linewidth]{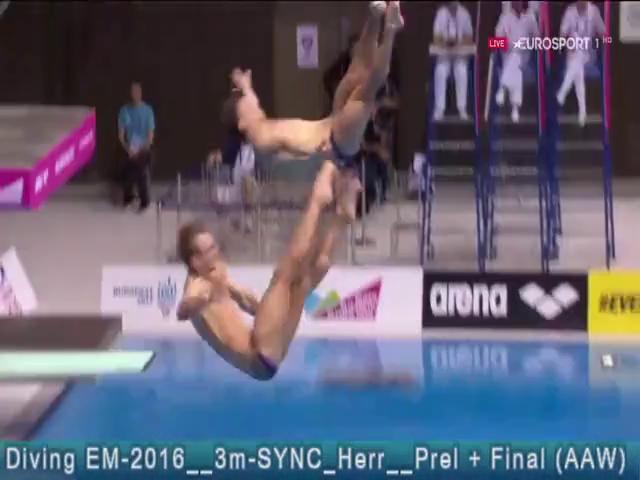}\\
    \includegraphics[width=0.12\linewidth]{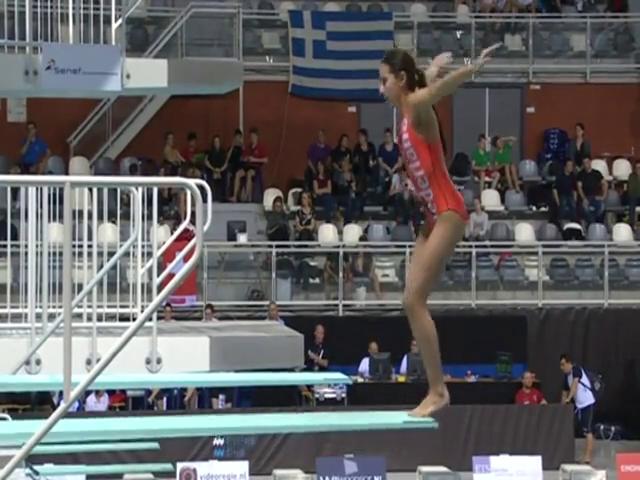}
    \includegraphics[width=0.12\linewidth]{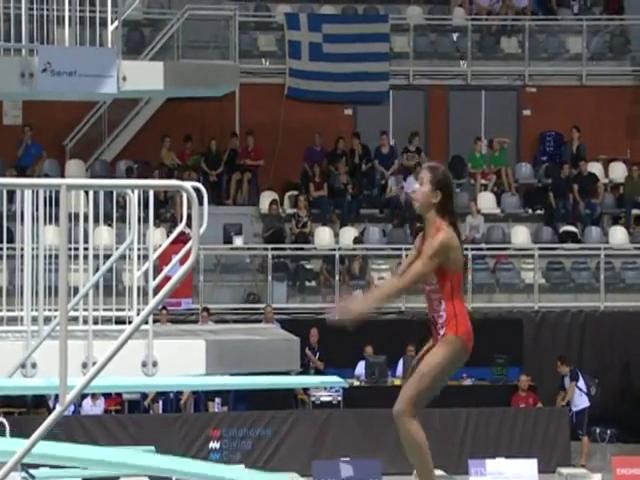}
    \includegraphics[width=0.12\linewidth]{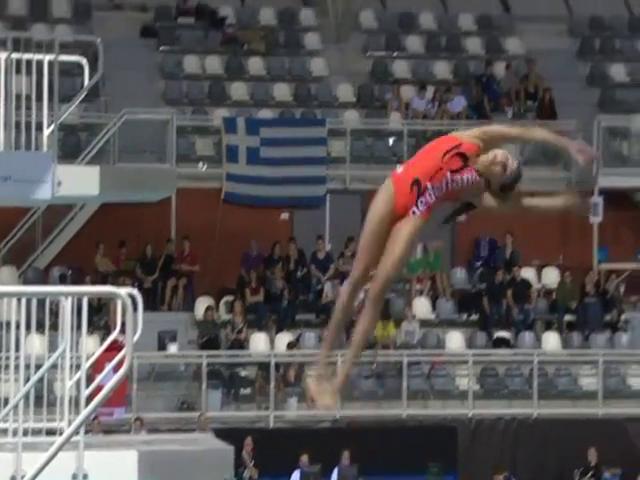}
    \includegraphics[width=0.12\linewidth]{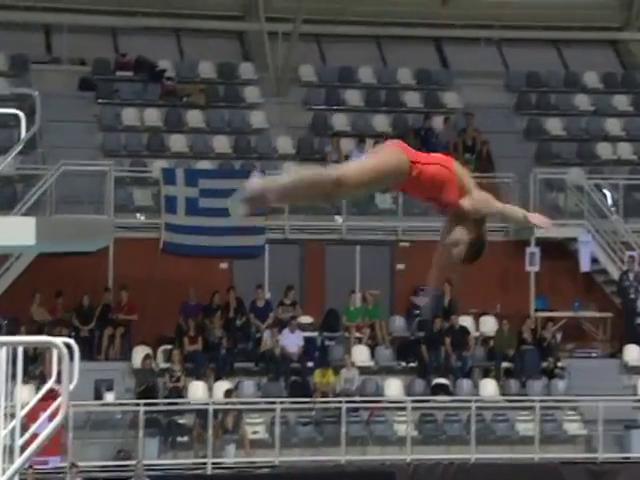}
    \includegraphics[width=0.12\linewidth]{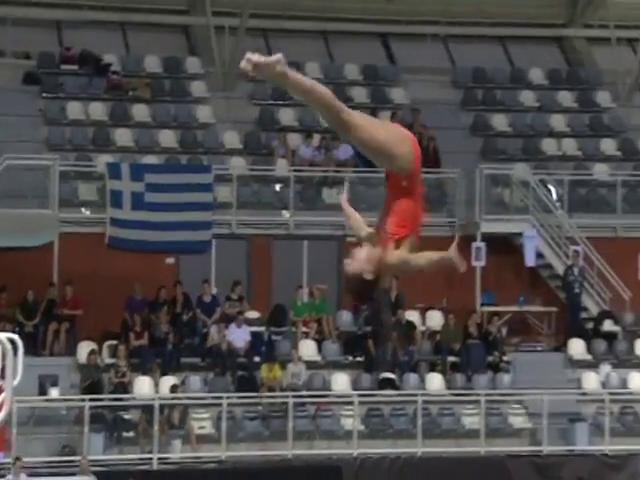}
    \includegraphics[width=0.12\linewidth]{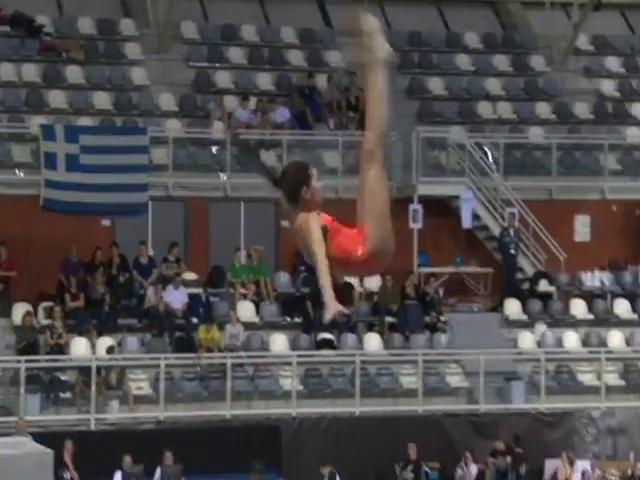}
    \includegraphics[width=0.12\linewidth]{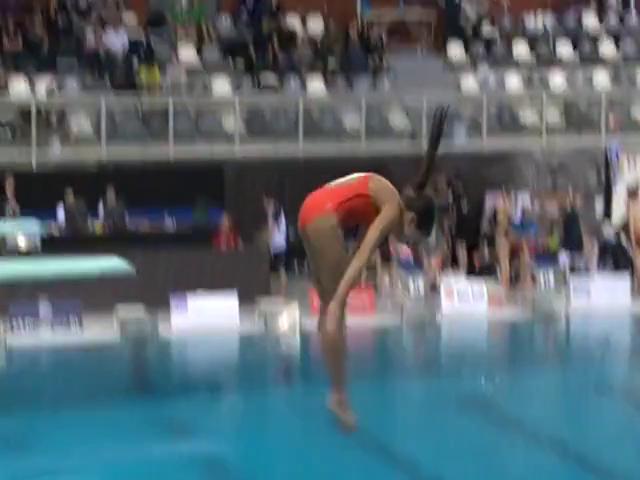}
    \includegraphics[width=0.12\linewidth]{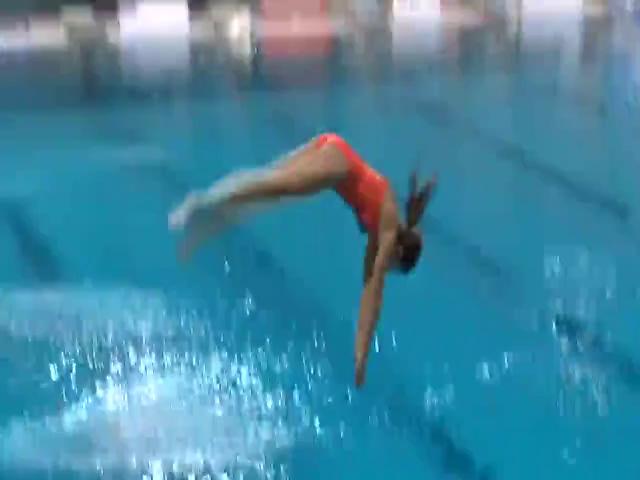}\\
    \includegraphics[width=0.12\linewidth]{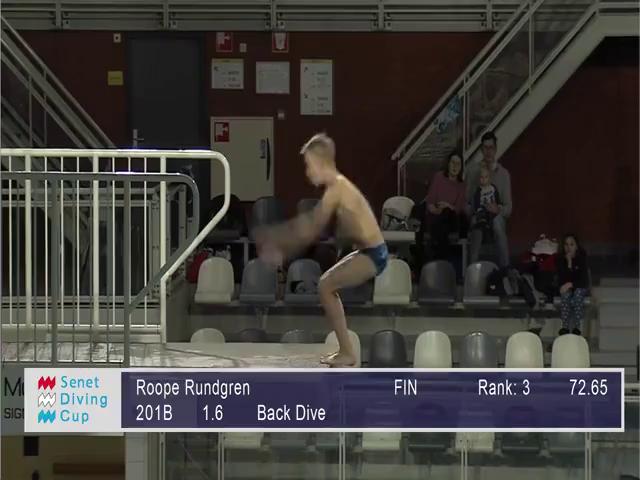}
    \includegraphics[width=0.12\linewidth]{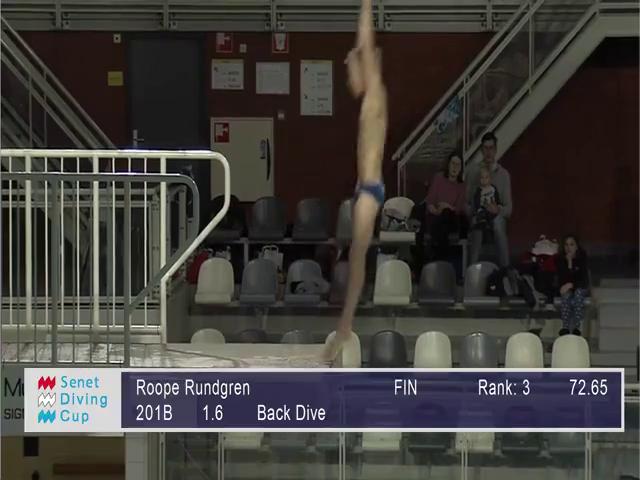}
    \includegraphics[width=0.12\linewidth]{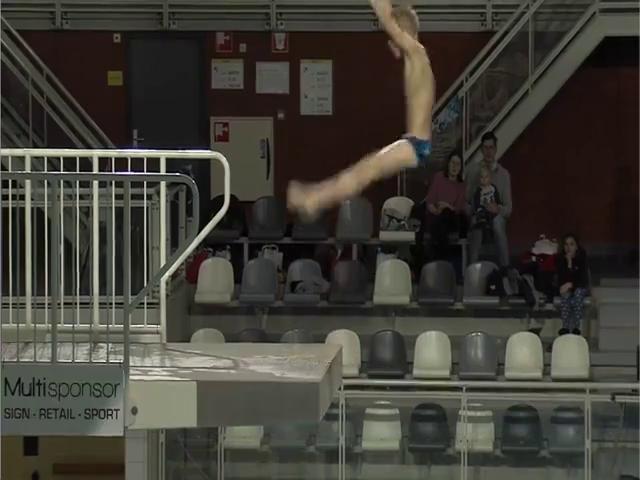}
    \includegraphics[width=0.12\linewidth]{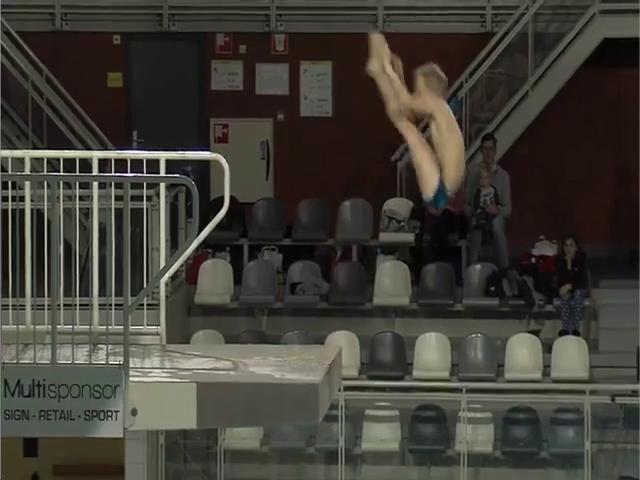}
    \includegraphics[width=0.12\linewidth]{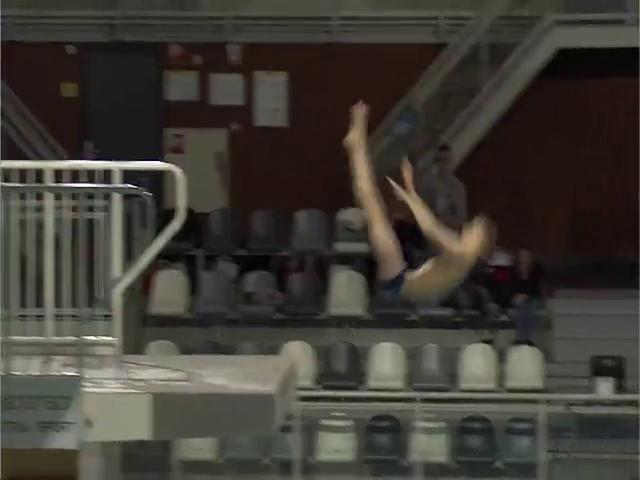}
    \includegraphics[width=0.12\linewidth]{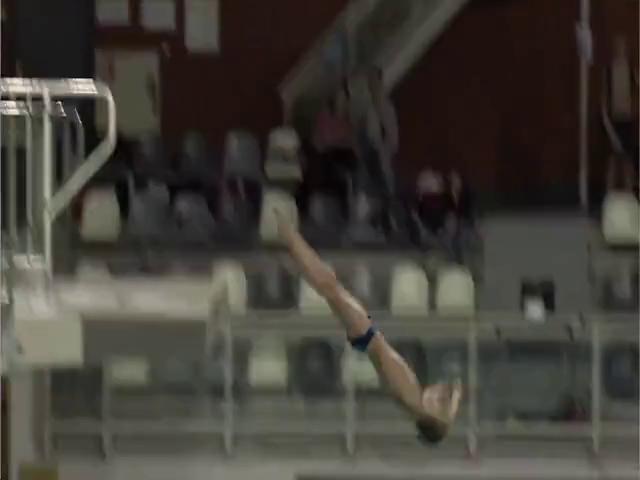}
    \includegraphics[width=0.12\linewidth]{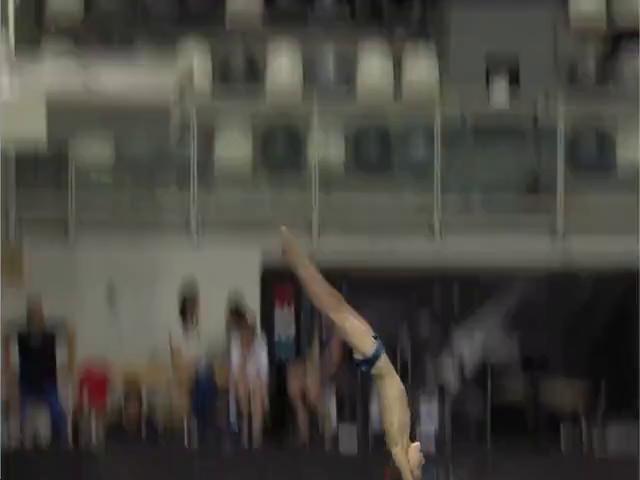}
    \includegraphics[width=0.12\linewidth]{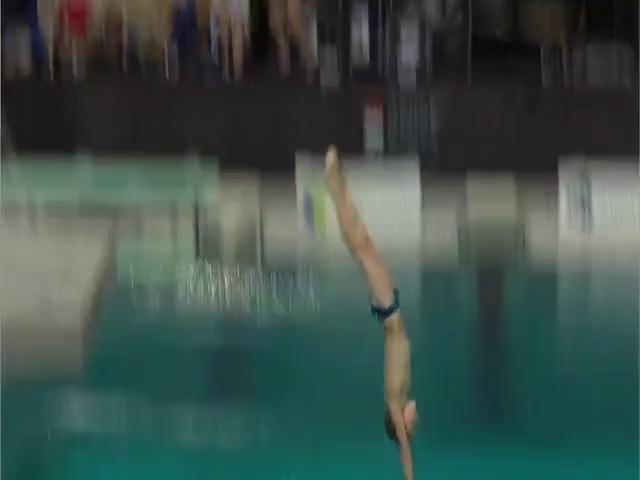}\\
    \includegraphics[width=0.12\linewidth]{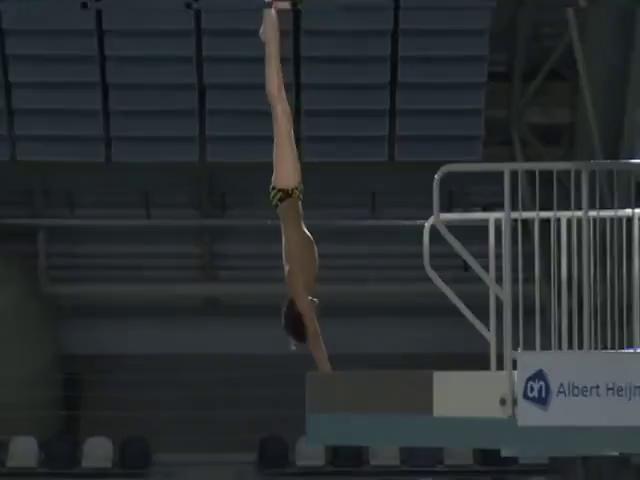}
    \includegraphics[width=0.12\linewidth]{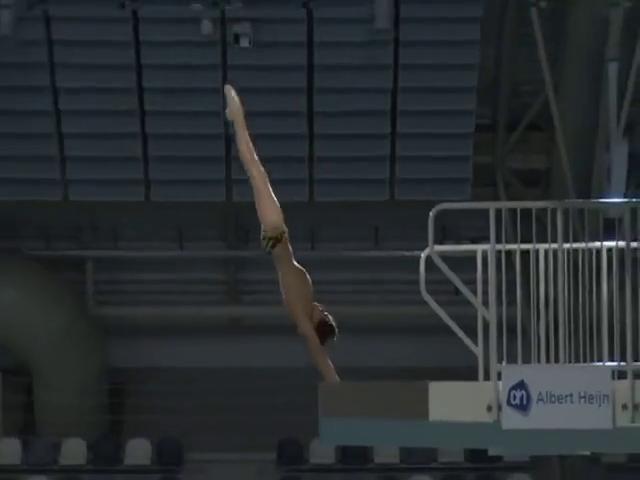}
    \includegraphics[width=0.12\linewidth]{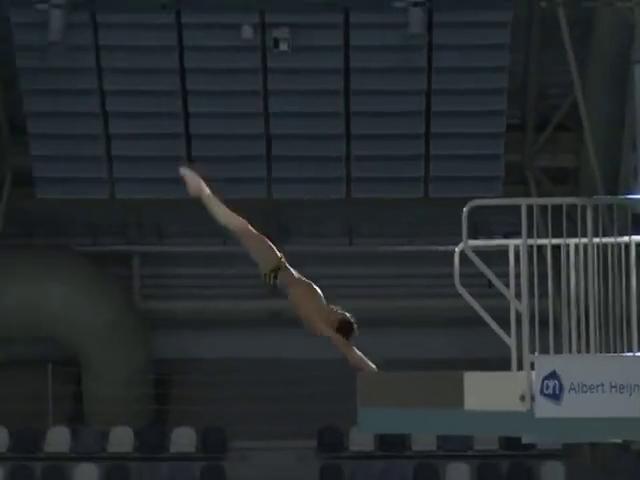}
    \includegraphics[width=0.12\linewidth]{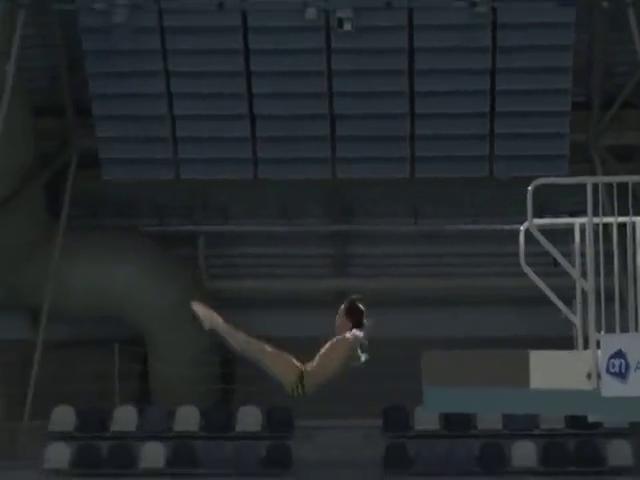}
    \includegraphics[width=0.12\linewidth]{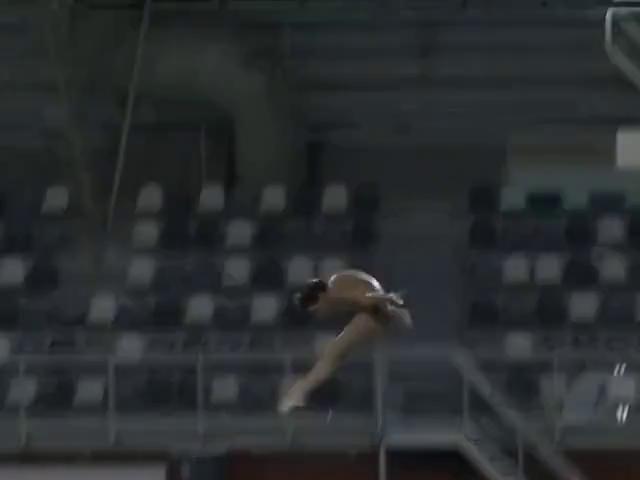}
    \includegraphics[width=0.12\linewidth]{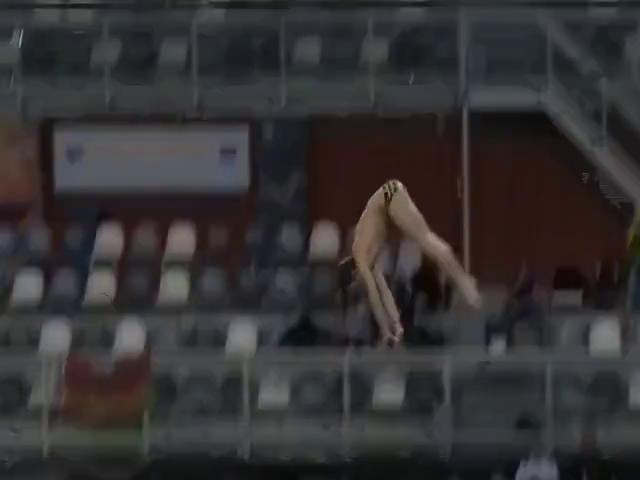}
    \includegraphics[width=0.12\linewidth]{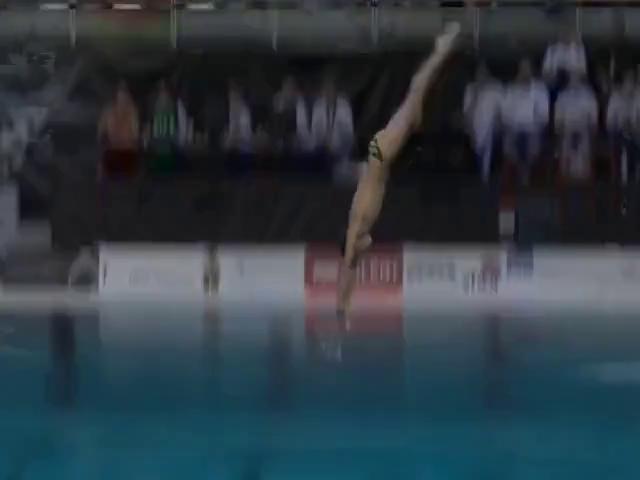}
    \includegraphics[width=0.12\linewidth]{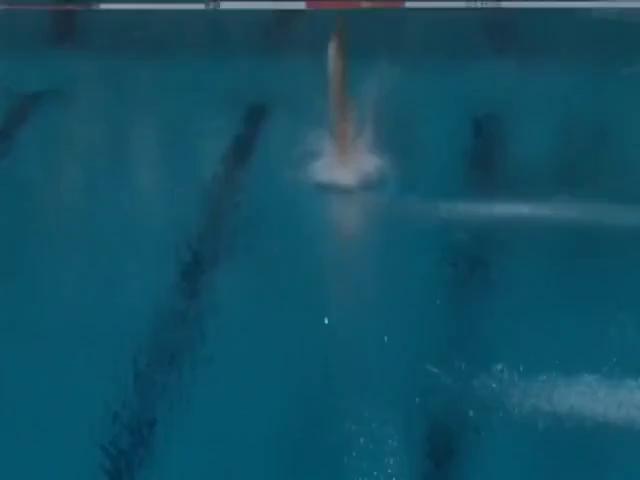}\\
    \includegraphics[width=0.12\linewidth]{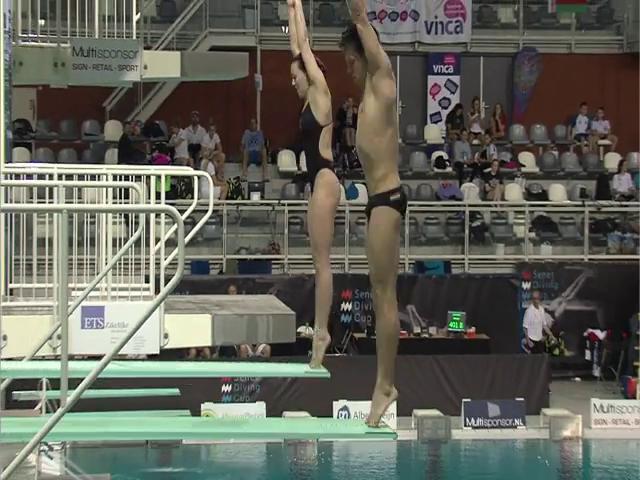}
    \includegraphics[width=0.12\linewidth]{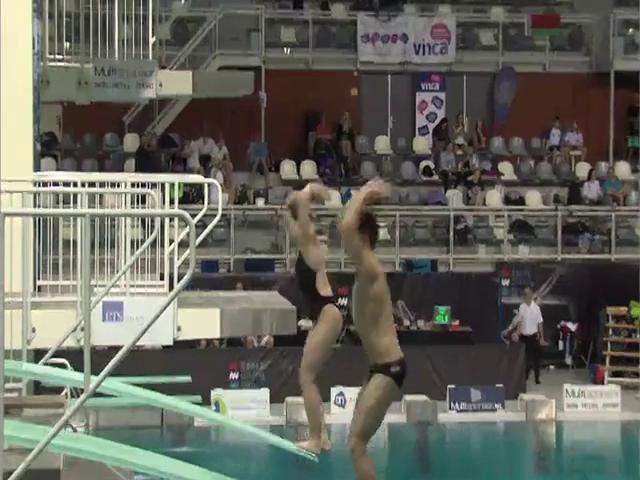}
    \includegraphics[width=0.12\linewidth]{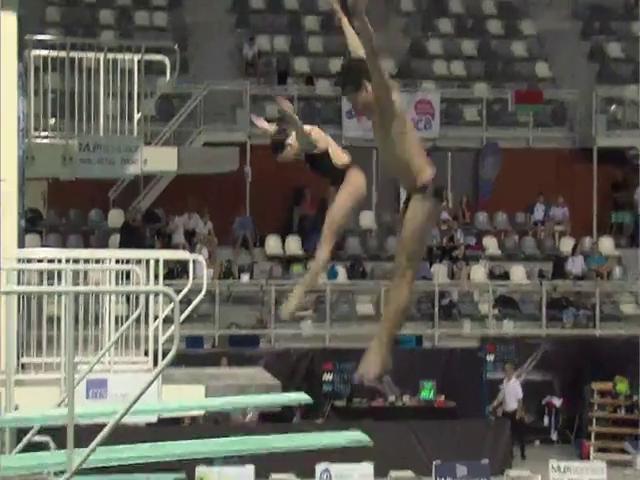}
    \includegraphics[width=0.12\linewidth]{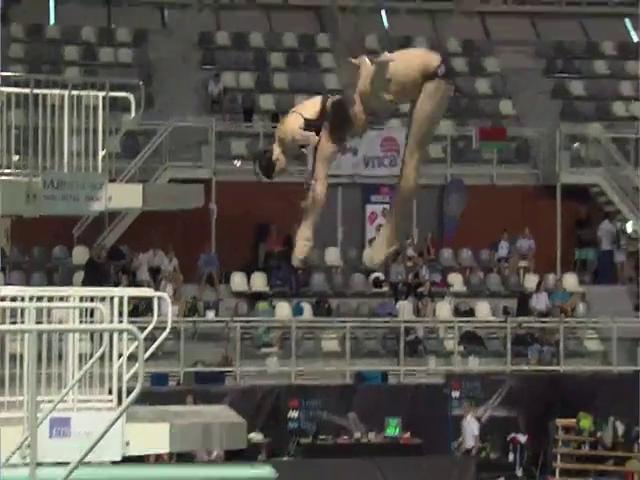}
    \includegraphics[width=0.12\linewidth]{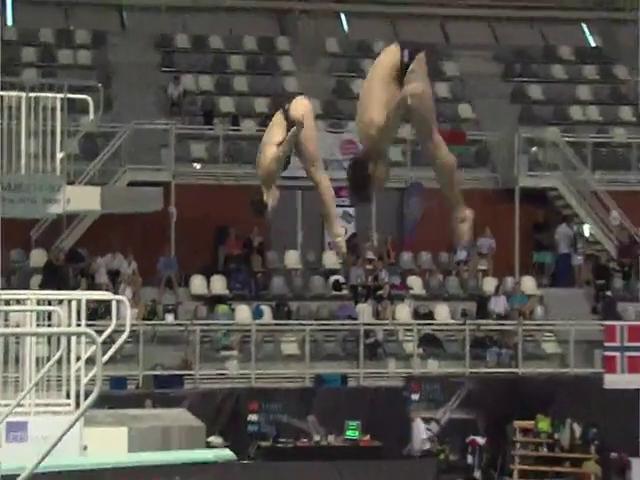}
    \includegraphics[width=0.12\linewidth]{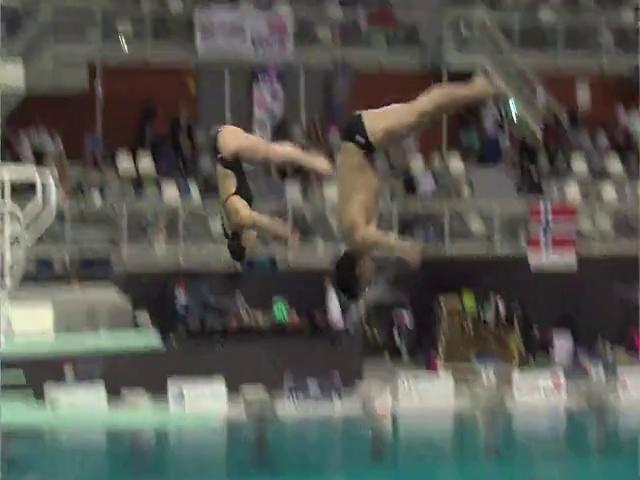}
    \includegraphics[width=0.12\linewidth]{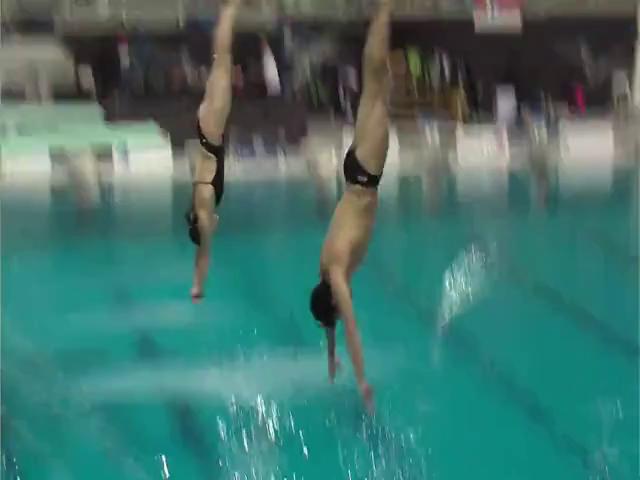}
    \includegraphics[width=0.12\linewidth]{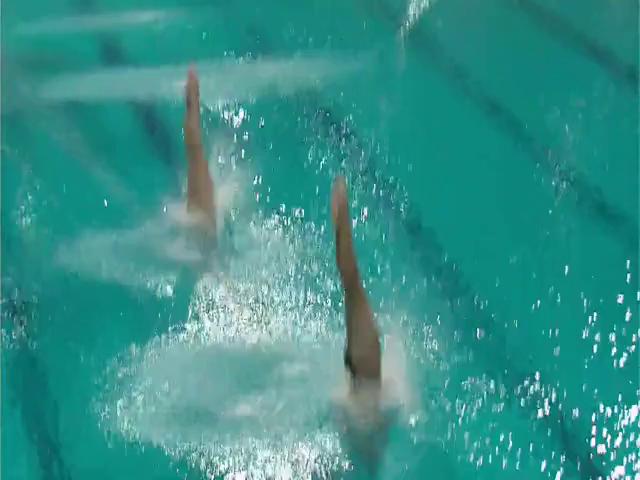}
    \caption{\textbf{Diving48 dataset}. Each row shows sample frames of the videos from the Diving48 dataset. The dataset contains 48 classes of diving. Each of the 48 diving class comprises four attributes: takeoff, somersaults, twists, and dive positions which have 4, 8, 8, and 4 different categories, respectively. It can be seen that the acrobatics performed by the divers are fairly complicated. It can also be noted that to classify each video to its class, the network should observe the diver through out the video. This brings about the need of having fine-grained spatio-temporal representations to capture the dynamics of the scene.}
    \label{fig:diving_samples}
\end{figure*}
In recent years, automatic interpretation of sports has gained a keen interest. It is a challenging task especially when it involves rapid changes and long-term dynamics. The increase in the processing power and accessibility to the huge data through broadcasting on digital media has enabled computer vision and machine learning techniques to perform several interesting and important tasks in this area such as object detection, action recognition, and player tracking.\\
In this work, we focus on competitive diving, which is a very popular sport in Olympics. In diving, a person jumps from a diving platform or a springboard and then performs some acrobatics before descending into the water. Based on the acrobatics performed during the dive, diving is classified into a finite set of action classes which are unambiguous in nature. These classes are standardized by FINA \cite{fina}. The differences in the acrobatics performed in various classes of diving are very subtle. The difference arises in the duration which starts with the diver standing on a diving platform or a springboard and ends at the moment he/she dives into the water. This brings about the need of having fine-grained spatio-temporal representations to capture the subtle differences between the different classes of  dives. Also, these representations should capture the long-term dynamics of the video for the task of diving classification.\\
In recent years, neural network based approaches have shown very promising results in the task of action recognition. This success is attributed to the introduction of several large datasets like UCF101 \cite{soomro2012ucf101}, HMDB51 \cite{kuehne2011hmdb}, PoseTrack \cite{Iqbal_CVPR2017}, Sports1M \cite{karpathy2014large}, and Kinetics \cite{kay2017kinetics}. These datasets contain videos of several types of actions. The networks are trained on these datasets to learn the task of action recognition. However, these datasets have strong static bias \cite{li2018resound} which lets the network perform well through learning static representations while making the dynamics present in the scene less meaningful for action recognition.\\
Recently, Li \emph{et al.} introduced a new dataset, Diving48, which contains over 18000 video clips of competitive diving actions \cite{li2018resound}. It contains 48 classes of diving. They defined each of the 48 diving classes by a combination of four attributes: takeoff, somersaults, twists, and dive positions which have $4,8,8,$ and $4$ different categories, respectively. They compared the representation biases of Diving48 with the existing datasets. They showed that the Diving48 has much smaller representation bias in comparison to several standard datasets. This brings about the need of capturing the scene dynamics in a proficient manner in order to perform well on this dataset. Figure \ref{fig:diving_samples} shows a few examples of diving videos from the Diving48 dataset. It can be seen that the moves are really complex and vary over the time of flight which makes it necessary to capture the fine-grained representations in order to capture those subtle movements along with the long-term dynamics.\\
 Long short-term memory (LSTM) is a variant of recurrent neural networks. It overcomes the problem of vanishing/exploding gradients in recurrent neural networks and is proven to be very successful in several computer vision and language processing tasks which involves long-term dependencies. They can accumulate the information for a long period of time. However, over a long period, several redundant information can get accumulated. LSTMs are well equipped in forgetting the redundant information and learning new information from the inputs at each time step. \\
 In this work, we propose an attention guided LSTM-based neural network for the task of diving classification. The proposed network is divided into four parts: feature extractor, encoder, attention network, and decoder. In diving, the classification is based on the actions performed by the diver. Hence, in this case, not all the spatial locations in the video frames are required for the classification. However, for the network to decide which spatial locations are necessary in each frame of the video, it requires a global context.  The role of the encoder network is to encode all the feature vectors extracted by the feature extractor from the frames of the videos to provide a global context to the attention and the decoder networks. Then, the attention network uses the global context to the generate the attention vectors for each frame which allows the classifier network to avoid any redundant information. Finally, the decoder network utilizes the global context and the feature vectors from each image weighted by the attention weights to obtain the classification results.\\
 The major contributions of this work are as follows.
 \begin{itemize}
     \item We propose a novel attention-guided LSTM-based neural network architecture for the task of diving classification.
     \item We show that the attention network learns to focus on the diver during the dive without being trained with such supervision.
     \item We experimentally show that the network learns better when the inputs to the attention network and the decoder network are given in the reverse order as compared to the order in which inputs are given to the encoder.
     \item The proposed network outperforms the state-of-the-art methods on the Diving48 dataset by 11.54\% and 4.24\% classification accuracy among 2D and  3D networks, respectively.
 \end{itemize}
The remaining part of the paper is organized as follows. Section \ref{sec:Related_Work} covers the literature survey of the relevant works. Section \ref{sec:Proposed_Approach} discusses the proposed network architecture for the task of diving classification. It also discusses the training procedure with the implementations details. Section \ref{sec:Experiments} discusses the experiments and the comparisons performed on the Diving48 dataset. Finally, Section \ref{sec:Conclusion} provides the conclusion for this work.
\section{Related Work}
\label{sec:Related_Work}
Computer vision along with machine learning has recently started to play an important role in sports. Its algorithms have a huge potential in many aspects of sports ranging from activity recognition, motion analysis of cameras and players, tracking players and objects, automatic annotation of sports footage, event detection, analysis of player injuries, spectator monitoring, performance assessment, and enhanced viewing. Here, we describe some of the recent works that use computer vision and machine learning for sports related tasks.
Shukla \emph{et al.} introduced a model that uses both event-based and excitement-based features to extract important events and cues from video shots in a cricket match in order to automatically generate match highlights \cite{shukla2018automatic}.  Li \emph{et al.} introduced a semi-supervised spatial transformer network for recognizing jersey number of the players  in soccer matches \cite{li2018jersey}. Their two stage model first detects the players on the court and then recognize their jersey number using a convolutional neural network.\\
Huda \emph{et al.} proposed a novel method for estimating the number of players in a soccer match  using simulation-based occlusion handling \cite{ul2018estimating}. They use a bagged tree classifier to first classify occlusion conditions. The number of players are then estimated by using the maximum likelihood of probability density based approach by further classifying the occluded players. Ullah and Cheikh in \cite{ullah2018directed} proposed a Directed Sparse Graphical Model (DSGM) for multi-target tracking in football, basketball, and sprint videos. The DGSM model finds a set of reliable tracks for the targets without any heuristics and keeps the computational complexity very low through the design of the graph. Reno \emph{et al.} introduced a CNN architecture for detecting ball in tennis games \cite{reno2018convolutional}.\\
Hwang \emph{et al.} proposed a model that combines global and local information for athlete pose estimation \cite{hwang2017athlete}. They first extract global information using a ResNet-101 based global network that is trained to regress a heat map representing parts' locations. The output features from the global network are then passed to the local network which learns spatial information using position sensitive score maps. Fani \emph{et al.} proposed an Integrated Stacked Hourglass Network (ISHN) for recognizing player actions in hockey videos \cite{fani2017hockey}. The ISHN network has three components. The first component is the latent pose estimator, the second transforms latent features to a common frame of reference, and the third performs action recognition.\\
Now, we discuss some of the important LSTM based works for activity recognition and trajectory prediction. Donahue \emph{et al.} proposed long-term recurrent convolutional networks  for activity recognition for learning long-term dependencies \cite{donahue2015long}. Alahi \emph{et al.} introduced social LSTM for predicting human trajectories in crowded spaces \cite{alahi2016social}. They viewed the problem of trajectory prediction as a sequence generation task, where interest is in predicting the future trajectory of people based on their past positions. Liu \emph{et al.} proposed a spatio-temporal LSTM based model with trust gates for 3D human action recognition \cite{liu2016spatio}. Veeriah \emph{et al.} proposed a differential recurrent neural network for action recognition \cite{veeriah2015differential}. Fragkiadaki \emph{et al.} proposed an encoder-recurrent-decoder (ERD)
model for recognition and prediction of human body pose in videos  \cite{fragkiadaki2015recurrent}.
\begin{figure*}[h!]
    \centering
    \includegraphics[width=0.83\linewidth]{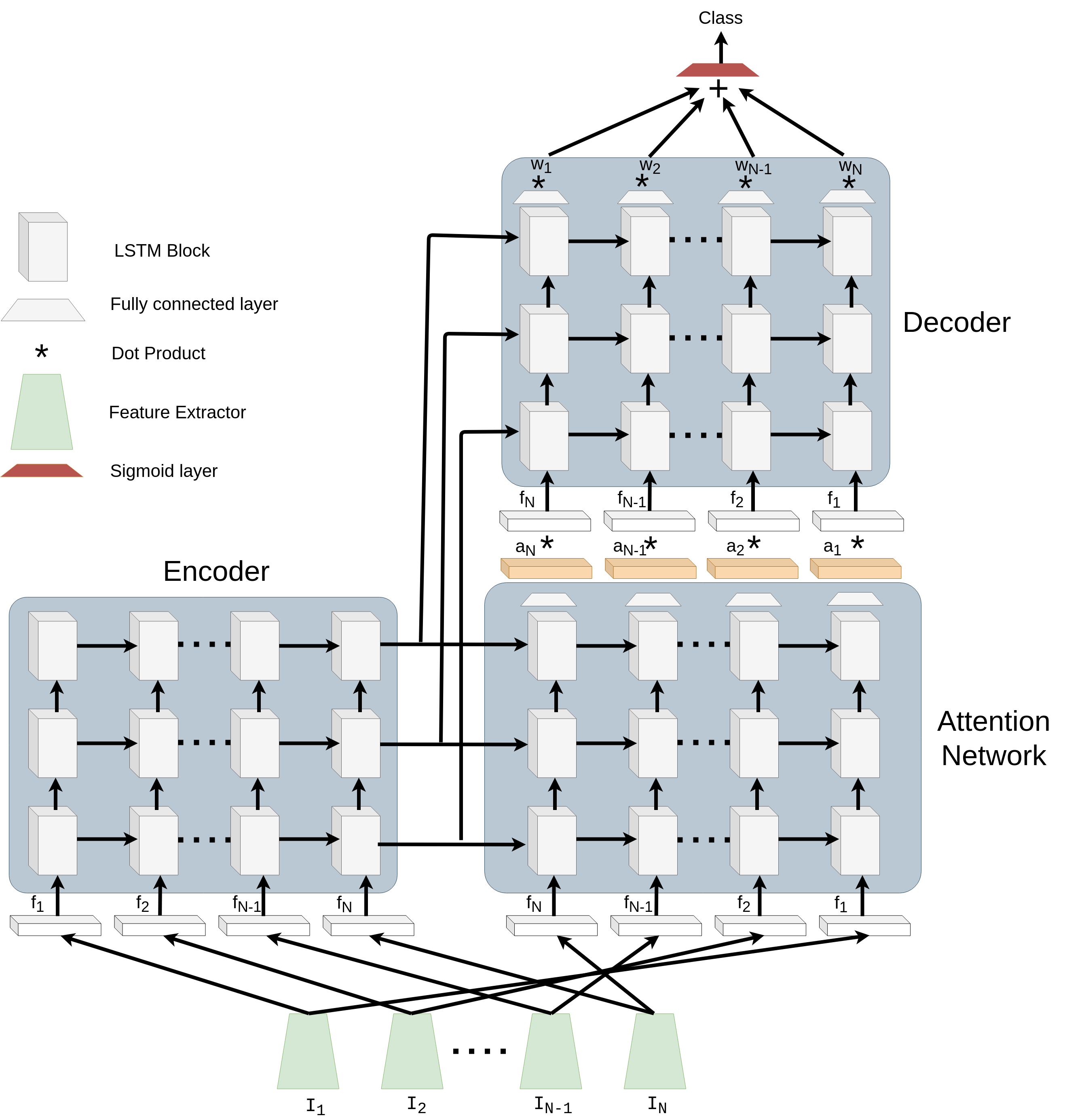}
    \caption{\textbf{The proposed network architecture.} The proposed model consists of four components: feature extractor, encoder, attention network, and decoder. $I_1$, $I_2$, \ldots, $I_N$ are the video frames of the input video which are fed to the feature extractor. $f_1$, $f_2$, \ldots, $f_N$ are the features extracted from the $N$ video frames using the feature extractor. We use ResNet (18 layers) as the feature extractor \cite{he2016deep}. The encoder, the attention network, and the decoder are multi-layer LSTM networks. The encoder takes $f_1$, $f_2$, \ldots, $f_N$ as input, one at each time step. The attention network takes $f_1$, $f_2$, \ldots, $f_N$ as inputs in the reverse order as compared to the encoder and generates the corresponding attention vectors $a_1$, $a_2$, \ldots, $a_N$. The decoder takes the dot-product of the attention vectors with their corresponding feature vector as input in the same order as of the attention network. The decoder outputs the representations corresponding to each video frame which are passed through a fully connected layer which are further used to perform the classification as shown in Eq.~\ref{Eq:classify}. $w_1$, $w_2$, \ldots $w_N$ are the learnable parameters.}
    \label{fig:Model}
\end{figure*}
\begin{figure*}[h]
    \centering
    \begin{minipage}{\linewidth}
    \centering
    \includegraphics[width=0.16\linewidth]{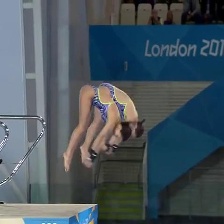}
    \includegraphics[width=0.16\linewidth]{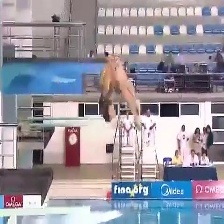}
    \includegraphics[width=0.16\linewidth]{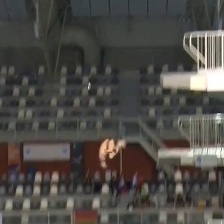}
    \includegraphics[width=0.16\linewidth]{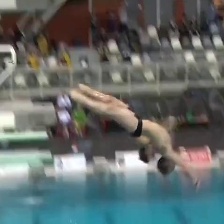}
    \includegraphics[width=0.16\linewidth]{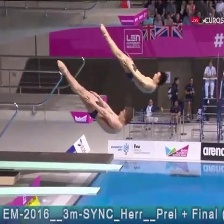}
    \includegraphics[width=0.16\linewidth]{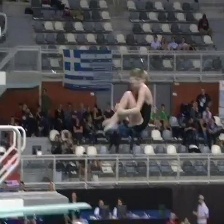}\\
    \includegraphics[width=0.16\linewidth]{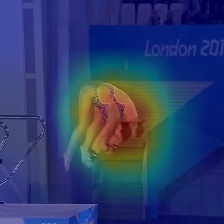}
    \includegraphics[width=0.16\linewidth]{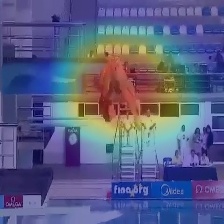}
    \includegraphics[width=0.16\linewidth]{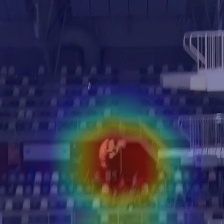}
    \includegraphics[width=0.16\linewidth]{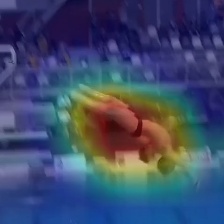}
    \includegraphics[width=0.16\linewidth]{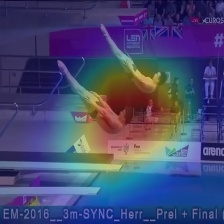}
    \includegraphics[width=0.16\linewidth]{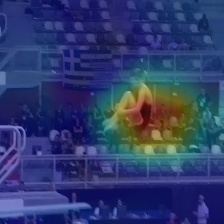}\\
    \includegraphics[width=0.16\linewidth]{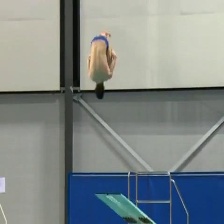}
    \includegraphics[width=0.16\linewidth]{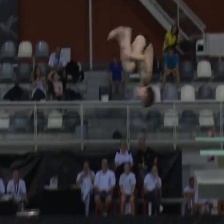}
    \includegraphics[width=0.16\linewidth]{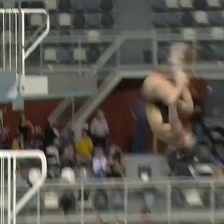}
    \includegraphics[width=0.16\linewidth]{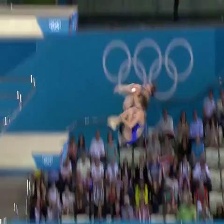}
    \includegraphics[width=0.16\linewidth]{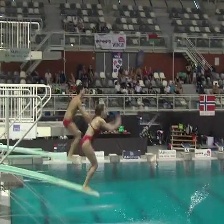}
    \includegraphics[width=0.16\linewidth]{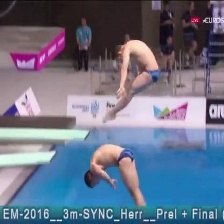}\\
    \includegraphics[width=0.16\linewidth]{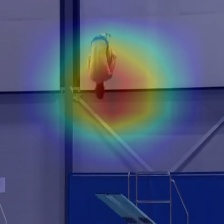}
    \includegraphics[width=0.16\linewidth]{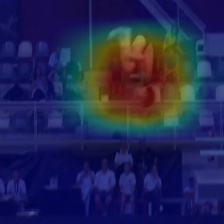}
    \includegraphics[width=0.16\linewidth]{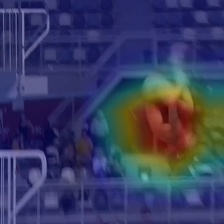}
    \includegraphics[width=0.16\linewidth]{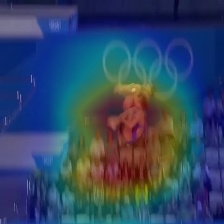}
    \includegraphics[width=0.16\linewidth]{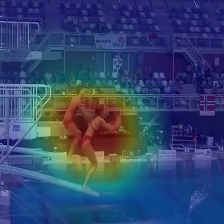}
    \includegraphics[width=0.16\linewidth]{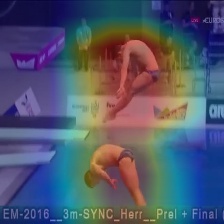}
    \end{minipage}
    \vspace{0.1em}
    \caption{\textbf{The attention network outputs}. The figure shows some examples of the attention obtained by the attention network on the Diving48 video frames. The first and the third row shows video frames of the videos from the test split of the Diving48 dataset. The second and the fourth row shows the attention map obtained by the attention network. It can be observed that the attention network is able to localize the diver without being with such supervision.}
    \label{fig:Atten_samples}
\end{figure*}
\section{Proposed Approach}
\label{sec:Proposed_Approach}
\subsection{Long Short-term Memory (LSTM) Network}
The LSTM was introduced by  Hochreiter and Schmidhuber \cite{hochreiter1997long}. The core of the LSTM is the cell state. It maintains its cell state by adding and removing the information from the cell state at each time step. LSTM has certain components, called gates, which helps the LSTM to maintain its cell state. When the LSTM receives an input, using the forget gate it decides what information to remove from the cell state and what to carry forward. Then, the input gate decides which values in the cell state should be updated and then an input modulation gate generates a vector which could be added to the cell state. Along with the cell state, hidden state is also propagated to the next time step, which is computed using the cell state at the current time step. LSTM performs the following equations to perform these tasks \cite{donahue2015long}.
\begin{equation}\label{eq:lstm_1}
i_t = \sigma(W_{xi}x_t+W_{hi}h_{t-1}+b_i)
\end{equation}
\begin{equation}\label{eq:lstm_2}
f_t = \sigma(W_{xf}x_t+W_{hf}h_{t-1}+b_f)
\end{equation}
\begin{equation}\label{eq:lstm_3}
o_t = \sigma(W_{xo}x_t+W^l_{ho}h_{t-1}+b_o)
\end{equation}
\begin{equation}\label{eq:lstm_4}
g_t = \tanh(W_{xc}x_t+W_{hc}h_{t-1}+b_c)
\end{equation}
\begin{equation}\label{eq:lstm_5}
c_t = f_t\odot c_{t-1}+i_t\odot g_t
\end{equation}
\begin{equation}\label{eq:lstm_6}
h_t = o_t \odot \tanh(c_t)
\end{equation}
Here, $\sigma$, $\tanh$ and $\odot$ stands for sigmoid function, tangent hyperbolic function, and dot product, respectively. $c_t  \in \mathbb{R}^p$ is the cell state, $h_t \in \mathbb{R}^p$ is the hidden unit, $i_t \in \mathbb{R}^p$ is the input gate, $f_t \in \mathbb{R}^p$ is the forget gate, $o_t  \in \mathbb{R}^p$ is the output gate, and $g_t \in \mathbb{R}^p$ is the input modulation gate. $W_{xi},W_{hi},W_{xf},W_{hf},W_{xo},W_{ho},W_{xc},W_{hc},b_i,b_f,b_0$, and $b_c$ are the learnable parameters. In this work, we have used a multi-layer LSTM in which multiple LSTMs are stacked in such a way that the input of the layer $l$ ($l>1$) is the hidden state of the previous layer.
\subsection{Model}
\label{Model}
The proposed model consists of four parts: feature extractor, encoder, attention network, and decoder.\\
\noindent\textbf{Feature extractor}.
We use a convolutional neural network as the feature extractor to obtain representations for each video frame. Let $\{I_i\}_{n=1}^N$ be the set of frames extracted from the input video. Here, $N$ is the number of frames extracted from the video. We pass $\{I_i\}_{n=1}^N$ through the feature extractor to obtain the feature vectors $\{f_i\}_{n=1}^N$ for each frame, where $f_i \in \mathbb{R}^m$, where $m$ is a constant. We use ResNet (18 layers) as the feature extractor for all the experiments \cite{he2016deep}. We remove the classification and global average pooling layer of ResNet (18 layers) \cite{he2016deep} and add a convolution layer of $1\times 1$ kernel size, stride of 1, and 1024 output feature channels. Then, we apply global average pooling and obtain 1024 dimensional feature vectors for each input video frame.\\
\noindent\textbf{Encoder}.
Encoder is a multi-layer LSTM network. The task of the encoder is to provide a global context of the input video frames to the attention network and the decoder. The encoder takes the feature vector $f_t$ as input at the time step $t$. The video frames are fed to the encoder in their temporal order. The encoder executes Eq.~\ref{eq:lstm_1}-~\ref{eq:lstm_6} for each $f_t$ at each time step $t$. We initialize the hidden and the cell state of the encoder with all zeros. The global context comprises the hidden and the cell state vectors obtained at the time step $t=N$.\\
\noindent\textbf{Attention network}.
Attention network is a multi-layer LSTM network followed by a fully connected layer. The attention network utilizes the global context provided by the encoder to generate the attention vectors for each feature representation $\{f_i\}_{n=1}^N$. We initialize the hidden and the cell state of the attention network with the hidden and the cell state vectors of encoder, respectively, obtained at the time step $t=N$. We feed the feature vectors $\{f_i\}_{n=1}^N$  to the attention network in the reverse order, i.e., first we feed $f_N$, then $f_{N-1}$, and so on \cite{sutskever2014sequence}. The intuition behind the reversing order is that the encoder has seen $f_N$ at the last time step. Hence, the attention network would be able to process $f_N$ in a much better way since its initial cell state is most familiar with $f_N$. Similarly, other feature vectors are given in the reverse temporal order. The attention network takes the feature vector $f_t$ at the time step $t$ and generates an output vector. This output vector is further passed through a fully connected layer to obtain the attention vector $a_t$, where $a_t \in \mathbb{R}^m$. Here, $m$ is the dimension of the outputs of the feature extractor.\\
\begin{table*}[t]   
	\centering
	\begin{tabular}[width=.9\textwidth]{l|c|c|c|c}
		\hline
		\hline
		\textbf{Network} & \textbf{Framework}	& \textbf{Input} & \textbf{Pre-training} & \textbf{Accuracy (\%)}  \\
		\hline	
		R(2+1)D \cite{tran2018closer, bertasius2018learning} & 3D & RGB & None & 21.4 \\
		C3D \cite{tran2015learning,li2018resound} & 3D & RGB & Sports1M (actions) & 27.6 \\
		R(2+1)D \cite{tran2018closer, bertasius2018learning} & 3D & RGB  & Kinetics (actions) & 28.9 \\
		Bertasius \emph{et al} \cite{bertasius2018learning} + R(2+1)D \cite{tran2018closer}  & 3D & RBG+Pose+DIMOFS & Kinetics + PoseTrack & 31.4 \\
		\hline
		TSN  \cite{wang2016temporal} & 2D &  RGB & ImageNet (objects)  & 16.8 \\
		Bertasius \emph{et al}  \cite{bertasius2018learning} & 2D & Pose & PoseTrack (poses) & 17.2 \\
		Bertasius \emph{et al} \cite{bertasius2018learning} & 2D & Pose+Flow &  PoseTrack (poses) & 18.8 \\
		TSN  \cite{wang2016temporal} & 2D & RGB+Flow& ImageNet (objects) & 20.3 \\
		TRN \cite{zhou2018temporal} & 2D & RGB+Flow & ImageNet (objects) & 22.8 \\
		Bertasius \emph{et al} \cite{bertasius2018learning} &  2D & Pose+DIMOFS & PoseTrack (poses) & 24.1 \\
		\hline
		Ours  & 2D & RGB & ImageNet (objects)  & \textbf{35.64} \\
		\hline
	\end{tabular}
	\vspace{1em}
	\caption{\textbf{Comparison with the state-of-the-art methods on the Diving48 dataset}. Our model significantly outperforms the state-of-the-art methods in both 2D and 3D frameworks. Note that our proposed network performs better than the models that have been pre-trained on larger-scale action recognition dataset such as Kinetics and uses just RGB images as input. DIMOFS stands for Discriminative Motion Features \cite{bertasius2018learning}.}
	\label{tab:results}
\end{table*}
\noindent\textbf{Decoder}.
Decoder is a multi-layer LSTM network. The decoder outputs the representations corresponding to each video frame which are used to obtain the probability of each diving class. Similar to the attention network, we initialize the hidden and the cell state of the decoder with the hidden and the cell state vectors of encoder, respectively, obtained at the time step $t=N$. The input $f^a_t$ to the decoder at the time step $t$ is computed as shown in Eq.~\ref{Eq:Decoder_input}.
\begin{equation}
    f^a_t = f_t \odot a_t
    \label{Eq:Decoder_input}
\end{equation}
Here, $f^a_t \in \mathbb{R}^m$ is the vector obtained by taking the dot-product of the feature vector $f_t$ with the attention vector $a_t$. Similar to attention network, we feed the vectors $\{f_n^a\}_{n=1}^N$ to the decoder in the reverse order. At each time step $t$, we feed $f^a_t$ to the decoder to obtain representation $f^r_t \in \mathbb{R}^p$ for $f^a_t$. The decoder outputs the set of representations $\{f_n^r\}_{n=1}^N$ which are passed through a fully connected layer $fc$ to obtain $\{\hat{o}_n\}_{n=1}^N$ which are further used to perform the classification as shown in Eq.~\ref{Eq:classify}.
\begin{equation}
    \hat{o} = \sum\limits_{n=1}^N w_n^r\hat{o}_n
    \label{Eq:classify}
\end{equation}
Here, $w_n^r \in \mathbb{R}$ is a learnable parameter, where $n=1,2,\ldots,N$. $\hat{o} \in \mathbb{R}^c$ is passed through a sigmoid layer to obtain the probability of each diving class.
\subsection{Training and Implementation Details}
\label{training}
\subsubsection{Dataset}
\begin{figure*}[h!]
    \centering
    \begin{minipage}{0.24\linewidth}
        \includegraphics[width = 0.49\linewidth]{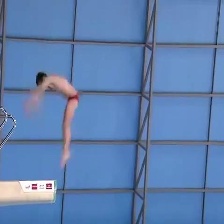}
        \includegraphics[width = 0.49\linewidth]{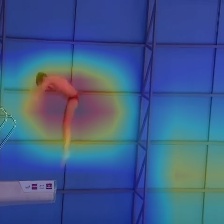}\\
        \includegraphics[width = 0.49\linewidth]{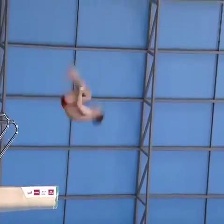}
        \includegraphics[width = 0.49\linewidth]{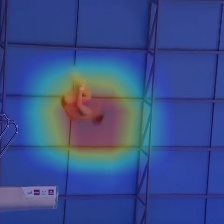}\\
        \includegraphics[width = 0.49\linewidth]{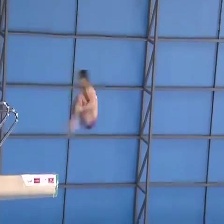}
        \includegraphics[width = 0.49\linewidth]{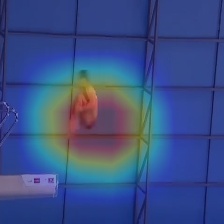}\\
        \includegraphics[width = 0.49\linewidth]{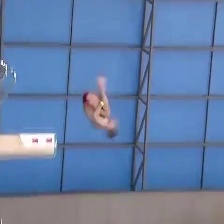}
        \includegraphics[width = 0.49\linewidth]{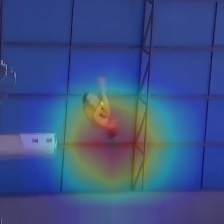}\\
        \includegraphics[width = 0.49\linewidth]{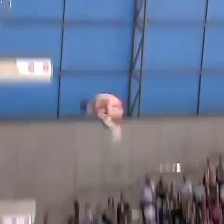}
        \includegraphics[width = 0.49\linewidth]{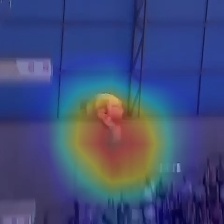}\\
        \includegraphics[width = 0.49\linewidth]{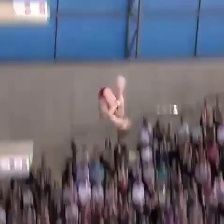}
        \includegraphics[width = 0.49\linewidth]{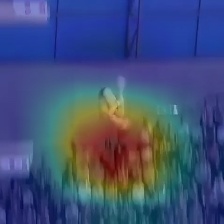}\\
        \includegraphics[width = 0.49\linewidth]{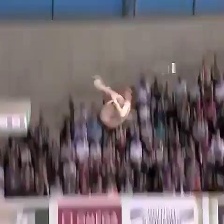}
        \includegraphics[width = 0.49\linewidth]{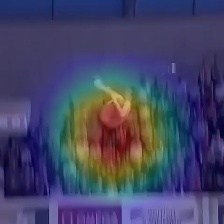}\\
        \includegraphics[width = 0.49\linewidth]{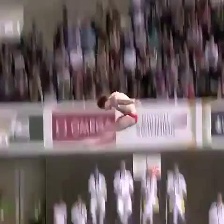}
        \includegraphics[width = 0.49\linewidth]{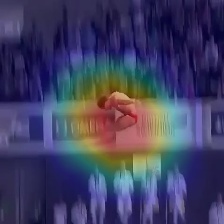}\\
        \includegraphics[width = 0.49\linewidth]{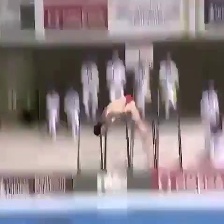}
        \includegraphics[width = 0.49\linewidth]{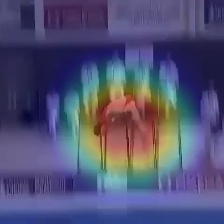}\\
        \includegraphics[width = 0.49\linewidth]{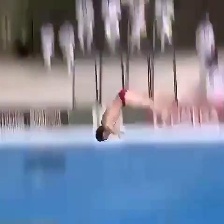}
        \includegraphics[width = 0.49\linewidth]{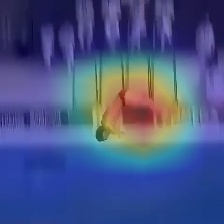}
    \end{minipage}
    \begin{minipage}{0.24\linewidth}
        \includegraphics[width = 0.49\linewidth]{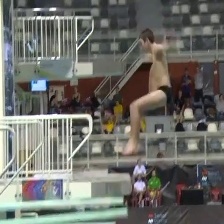}
        \includegraphics[width = 0.49\linewidth]{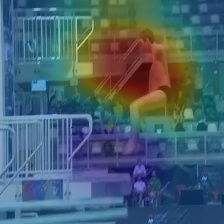}\\
        \includegraphics[width = 0.49\linewidth]{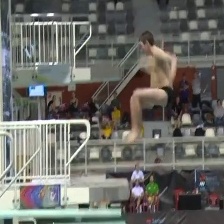}
        \includegraphics[width = 0.49\linewidth]{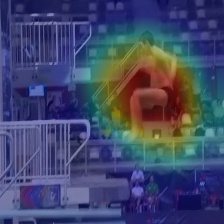}\\
        \includegraphics[width = 0.49\linewidth]{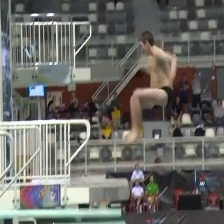}
        \includegraphics[width =0.49\linewidth]{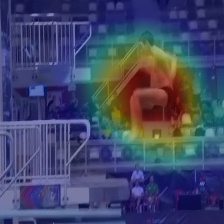}\\
        \includegraphics[width = 0.49\linewidth]{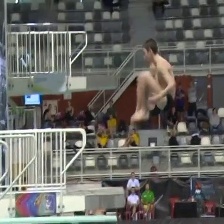}
        \includegraphics[width = 0.49\linewidth]{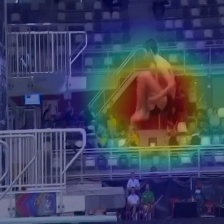}\\
        \includegraphics[width = 0.49\linewidth]{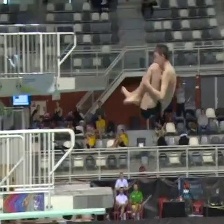}
        \includegraphics[width = 0.49\linewidth]{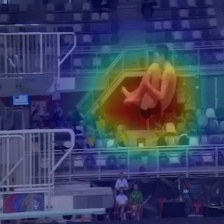}\\
        \includegraphics[width = 0.49\linewidth]{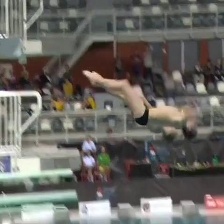}
        \includegraphics[width = 0.49\linewidth]{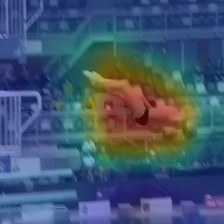}\\
        \includegraphics[width = 0.49\linewidth]{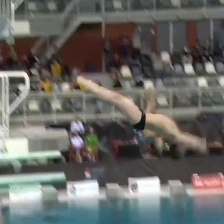}
        \includegraphics[width = 0.49\linewidth]{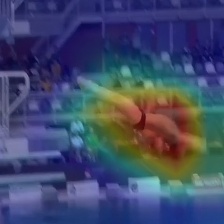}\\
        \includegraphics[width = 0.49\linewidth]{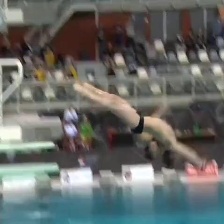}
        \includegraphics[width = 0.49\linewidth]{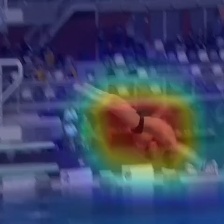}\\
        \includegraphics[width = 0.49\linewidth]{b_im_32.jpg}
        \includegraphics[width = 0.49\linewidth]{b_32.jpg}\\
        \includegraphics[width = 0.49\linewidth]{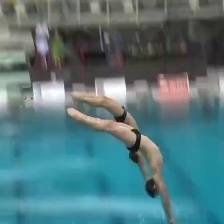}
        \includegraphics[width = 0.49\linewidth]{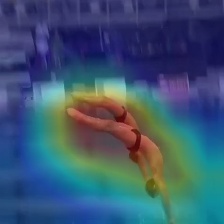}
    \end{minipage}
    \begin{minipage}{0.24\linewidth}
        \includegraphics[width = 0.49\linewidth]{c_im_00.jpg}
        \includegraphics[width = 0.49\linewidth]{c_00.jpg}\\
        \includegraphics[width = 0.49\linewidth]{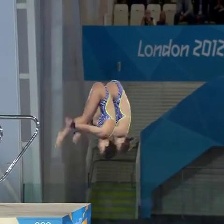}
        \includegraphics[width = 0.49\linewidth]{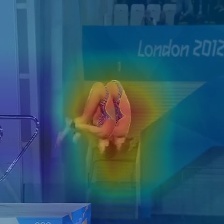}\\
        \includegraphics[width = 0.49\linewidth]{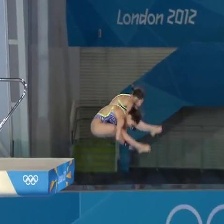}
        \includegraphics[width =0.49\linewidth]{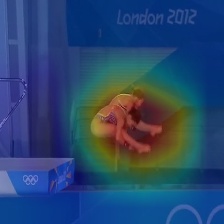}\\
        \includegraphics[width = 0.49\linewidth]{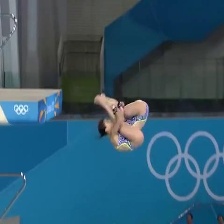}
        \includegraphics[width = 0.49\linewidth]{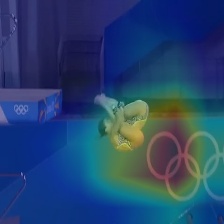}\\
        \includegraphics[width = 0.49\linewidth]{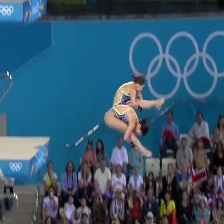}
        \includegraphics[width = 0.49\linewidth]{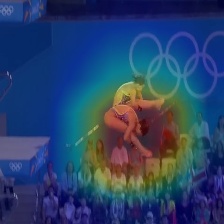}\\
        \includegraphics[width = 0.49\linewidth]{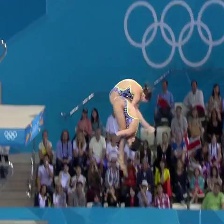}
        \includegraphics[width = 0.49\linewidth]{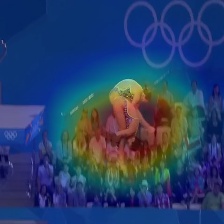}\\
        \includegraphics[width = 0.49\linewidth]{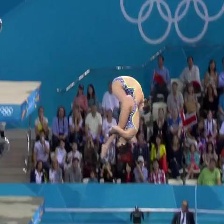}
        \includegraphics[width = 0.49\linewidth]{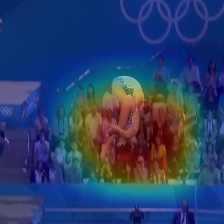}\\
        \includegraphics[width = 0.49\linewidth]{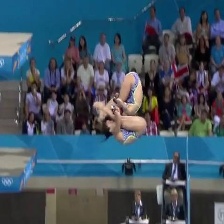}
        \includegraphics[width = 0.49\linewidth]{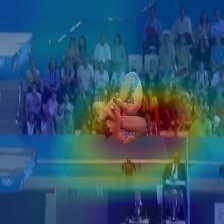}\\
        \includegraphics[width = 0.49\linewidth]{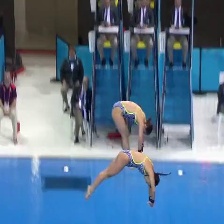}
        \includegraphics[width = 0.49\linewidth]{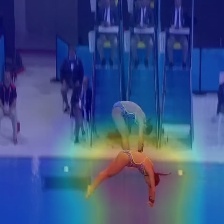}\\
        \includegraphics[width = 0.49\linewidth]{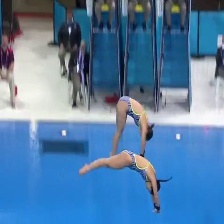}
        \includegraphics[width = 0.49\linewidth]{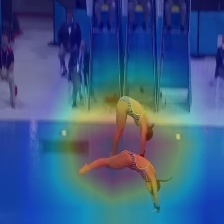}
    \end{minipage}
    \begin{minipage}{0.24\linewidth}
        \includegraphics[width = 0.49\linewidth]{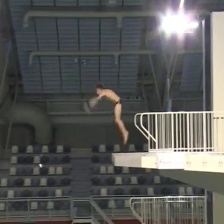}
        \includegraphics[width = 0.49\linewidth]{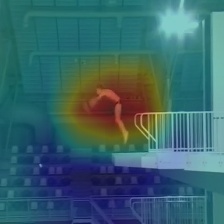}\\
        \includegraphics[width = 0.49\linewidth]{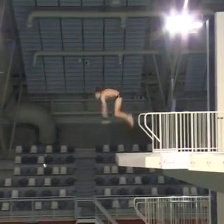}
        \includegraphics[width = 0.49\linewidth]{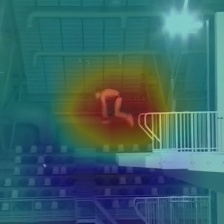}\\
        \includegraphics[width = 0.49\linewidth]{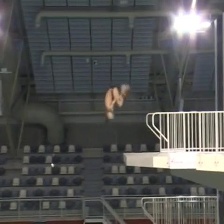}
        \includegraphics[width =0.49\linewidth]{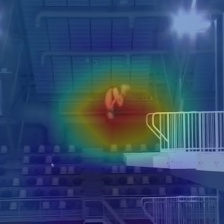}\\
        \includegraphics[width = 0.49\linewidth]{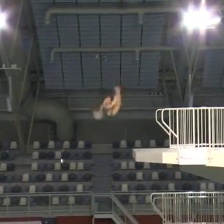}
        \includegraphics[width = 0.49\linewidth]{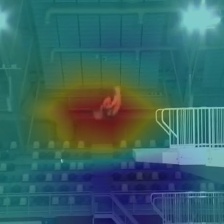}\\
        \includegraphics[width = 0.49\linewidth]{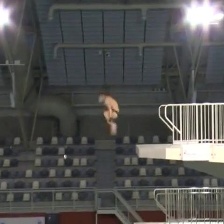}
        \includegraphics[width = 0.49\linewidth]{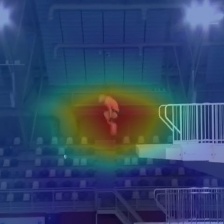}\\
        \includegraphics[width = 0.49\linewidth]{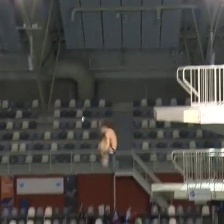}
        \includegraphics[width = 0.49\linewidth]{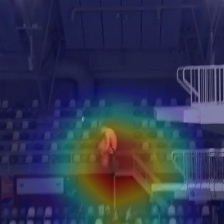}\\
        \includegraphics[width = 0.49\linewidth]{d_im_38.jpg}
        \includegraphics[width = 0.49\linewidth]{d_38.jpg}\\
        \includegraphics[width = 0.49\linewidth]{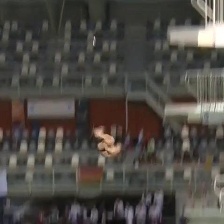}
        \includegraphics[width = 0.49\linewidth]{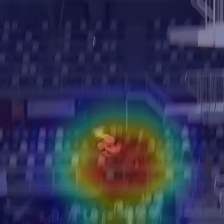}\\
        \includegraphics[width = 0.49\linewidth]{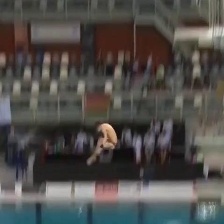}
        \includegraphics[width = 0.49\linewidth]{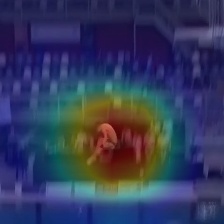}\\
        \includegraphics[width = 0.49\linewidth]{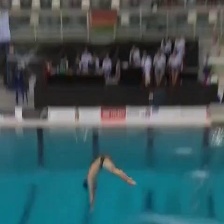}
        \includegraphics[width = 0.49\linewidth]{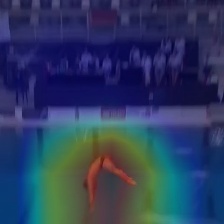}
    \end{minipage}
       \vspace{0.7em}
    \caption{\textbf{Attention visualization}. The figure shows visualization of the outputs of the attention network of our proposed approach on the video frames from the test split of Diving48 dataset. It can be seen that the attention network is able to localize the diver(s) in the video frames without being trained with such supervision}
    \label{fig:atten_results}
\end{figure*}
Diving48 is an action recognition dataset which contains 18,404 videos clips of competitive diving actions. The clips cover 48 diving classes which are standardized by FINA \cite{fina}. There are 16,067 clips in the training set and 2,337 clips in the test set. They represent each diving class as a sequence of four attributes: Take-off, Somersault, Twist, and Flight position which are divided into 4,8,8, and 4 classes, respectively. In the dataset, it is ensured that the clips from the same competition are assigned to the same set.
\subsubsection{Representation Learning}
Instead of directly training the network to identify the diving classes, we train the network to identify the four attributes of the diving classes. The output of the decoder is passed through four fully connected layers $fc_{1},fc_2,fc_3$, and $fc_4$ which predict the classes of the four attributes, i.e., Take-off, Somersault, Twist, and Flight position, respectively. Let $\{\hat{o}_n^1\}_{n=1}^N$, $\{\hat{o}_n^2\}_{n=1}^N$, $\{\hat{o}_n^3\}_{n=1}^N$, and $\{\hat{o}_n^4\}_{n=1}^N$ be the outputs of $fc_{1},fc_2,fc_3$, and $fc_4$ for $N$ input video frames. We train the network using the loss function shown in Eq. \ref{Eq:loss}.
\begin{equation}
    \mathcal{L}_r = \sum\limits_{i=1}^4 S\bigg(\sum\limits_{n=1}^N w_n^i\hat{o}_n^i\bigg)
    \label{Eq:loss}
\end{equation}
Here, $S$ is the cross-entropy loss and $w_n^i$ is a learnable parameter where $i=1,2,3$, and $4$ and $n=1,2,\ldots,N$. By training the network with the loss function $\mathcal{L}_r$, we learn the representations $\{f_n^r\}_{n=1}^N$, which are the output of the decoder.\\
We then remove the fully connected layers $fc_{1},fc_2,fc_3$, and $fc_4$, and add the fully connected layer $fc$ on the output of the decoder network. We keep the trained weights of the network and only train $fc$ to predict the diving class using the loss function $S(\hat{o})$. Here, $S$ is the cross-entropy loss and $\hat{o}$ is defined in Eq. \ref{Eq:classify}.\\
We use Adam for the weight update with the initial learning rate of $10^{-4}$, $\beta_1=0.9$, $\beta_2=0.999$, and $\epsilon= 10^{-8}$ \cite{kingma2014adam}. We apply a dropout of 0.2 on all the fully connected layers. We use 512 hidden units in the encoder and the attention networks and 256 hidden units in the decoder network. For the data augmentation, while training, we resize the shorter edge of each video frame of the video to 245 pixels and then, we randomly extract a volume of $224 \times 224 \times N$. The $N$ frames are picked randomly from the input video following a uniform distribution. We have used 64 frames of each video for training and testing the network, i.e., $N=64$. We implemented our network and performed all the experiments using PyTorch on a system with Intel i7-7820X processor, 32 GB RAM, and an Nvidia Titan Xp GPU.
\section{Experiments}
\label{sec:Experiments}
In this section, we evaluate the effectiveness of the proposed model on the Diving48 dataset. We also evaluate the effect of the design choices by performing multiple ablation studies. We show that the attention network is able to localize the diver in the video frames without being trained with such supervision.
\subsection{Results}
We compared our results on Diving48 datasets with the state-of-the-art methods. Table \ref{tab:results} shows the comparison of the proposed approach with the state-of-the-art methods. The proposed model obtains 35.64\% classification accuracy which outperforms the state-of-the-art by 11.54\% among 2D networks and by 4.24\% among 3D networks. We used pretrained weights of ResNet (18 layers) trained on ImageNet \cite{russakovsky2015imagenet} which is used as the feature extractor in the proposed model. The encoder, the attention and the decoder networks are trained from scratch on the Diving48 dataset. Figure~\ref{fig:atten_results} shows the attention maps obtained using the Eq.~\ref{Eq:Attention_maps}.
\begin{equation}
    \mathcal{A}_t = \sum\limits_i a_t^iF_i
    \label{Eq:Attention_maps}
\end{equation}
Here, $\mathcal{A}_t$ is the attention map for the video frame which is provided as the input to the decoder at time step $t$, $a_t = (a_t^1,a_t^2,\ldots,a_t^m)$ is the output vector of the attention network and $\{F_i\}_{i=1}^m$ are the feature maps of the corresponding video frame obtained from the last layer of the feature extractor \cite{zhou2015cnnlocalization}. We upsample $\mathcal{A}_t$ to the size of the corresponding input image to overlay it. In Figure \ref{fig:atten_results}, it can be seen that the network is not only able to localize the diver but also able to follow it through the video without being trained for such a task. The network is observed to ignore the static parts of the scene and is also observed to focus on the divers which is actually relevant to the task.
\subsection{Ablation Studies}
We perform the following ablation studies to quantify the effect of choice of components for the network.\\
\noindent\textbf{Attention Network}.
We removed the attention network and trained the remaining networks in the same manner as the proposed model. In this case, the input of the decoder is the feature vector $f_t$, instead of vector $f_t^a$ which is obtained by taking the dot product of the attention vector with $f_t$. Table \ref{tab:without_attention} shows the results obtained by training the proposed model without the attention network. It can be seen that there is a drastic decrease in the accuracy which shows its importance in the proposed model.\\
\begin{table}[t]   
	\centering
	\begin{tabular}[width=.9\textwidth]{l|c}
		\hline
		\hline
		\textbf{Network}  & \textbf{Accuracy (\%)}  \\
		\hline	
		  Without Attention   &  29.63\\
		  With Attention   & \textbf{35.64} \\
		\hline
	\end{tabular}
	\vspace{1em}
	\caption{\textbf{Clip accuracy with and without the attention network on the test split of Diving48 dataset}. The proposed model with the attention network performs significantly better. }
	\label{tab:without_attention}
\end{table}
\noindent\textbf{Representation Learning}.
In this study, we checked the necessity of first learning the representations using the attributes of diving and then learning the diving classification. Instead of first learning the representations, we directly trained the proposed model to learn to classify the diving videos into its classes. Table \ref{tab:without_representation} shows the comparison of the classification accuracy with and without representation learning. With representation learning, we obtain better accuracy which experimentally proves its usefulness.
\begin{table}[t]   
	\centering
	\begin{tabular}[width=.9\textwidth]{l|c}
		\hline
		\hline
		\textbf{Network}  & \textbf{Accuracy (\%)}  \\
		\hline	
		  Without Representation  & 32.54 \\
		  With Representation  & \textbf{35.64} \\
		\hline
	\end{tabular}
	\vspace{1em}
	\caption{\textbf{Clip accuracy with and without representation learning on the test split of Diving48 dataset}. The proposed model performs better when first trained on class attributes and then fine-tuned on the diving classes.}
	\label{tab:without_representation}
\end{table}
\noindent\textbf{Unreversed sequence}.
In Section \ref{Model}, we have mentioned that the feature vectors $\{f_i\}_{n=1}^N$ and $\{f_n^a\}_{n=1}^N$ are fed to the attention and the decoder networks, respectively, in the reversed order as compared to sequence in which $\{f_i\}_{n=1}^N$ are fed to the encoder network. In this study, we evaluate the effect of reversing the sequence. We train the network by feeding the feature vectors to the encoder, the attention network, and the decoder in the same sequence. Table \ref{tab:without_representation} compares the accuracy when the decoder and the attention networks are trained with unreversed sequence with the reversed sequence case.
\begin{table}[t]   
	\centering
	\begin{tabular}[width=.9\textwidth]{l|c}
		\hline
		\hline
		\textbf{Network}  & \textbf{Accuracy (\%)}  \\
		\hline	
		  Sequence unreversed  & 32.59 \\
		  Sequence Reversed  & \textbf{35.64} \\
		\hline
	\end{tabular}
	\vspace{1em}
		\caption{\textbf{Clip accuracy on the test split of Diving48 dataset}. The proposed model performs better when decoder and attention networks are fed with reversed sequence.}
	\label{tab:without_representation}
\end{table}
\section{Conclusion}
\label{sec:Conclusion}
We propose a novel attention guided LSTM-based neural network architecture for diving classification. We evaluate the proposed approach on a standard competitive diving classification dataset, Diving48. The proposed model outperforms the state-of-the-art methods by a significant margin. We show that the network learns better when we first train it on the attributes of diving and then train the classification layer on the diving classes rather than directly training on the diving classes. We also show that the attention network localizes the diver in the diving videos without being trained with such supervision.\\

\noindent\textbf{Acknowledgments.} Gagan Kanojia and Sudhakar Kumawat were supported by TCS Research Fellowships. Shanmuganathan Raman was supported by SERB Core Research Grant and Imprint 2 Grant.


\begin{thebibliography}{10}\itemsep=-1pt

\bibitem{fina}
F´ed´eration internationale de natation.
\newblock http://www.fina.org/.

\bibitem{alahi2016social}
Alexandre Alahi, Kratarth Goel, Vignesh Ramanathan, Alexandre Robicquet, Li
  Fei-Fei, and Silvio Savarese.
\newblock Social lstm: Human trajectory prediction in crowded spaces.
\newblock In {\em Proceedings of the IEEE conference on computer vision and
  pattern recognition}, pages 961--971, 2016.

\bibitem{bertasius2018learning}
Gedas Bertasius, Christoph Feichtenhofer, Du Tran, Jianbo Shi, and Lorenzo
  Torresani.
\newblock Learning discriminative motion features through detection.
\newblock {\em arXiv preprint arXiv:1812.04172}, 2018.

\bibitem{donahue2015long}
Jeffrey Donahue, Lisa Anne~Hendricks, Sergio Guadarrama, Marcus Rohrbach,
  Subhashini Venugopalan, Kate Saenko, and Trevor Darrell.
\newblock Long-term recurrent convolutional networks for visual recognition and
  description.
\newblock In {\em Proceedings of the IEEE conference on computer vision and
  pattern recognition}, pages 2625--2634, 2015.

\bibitem{fani2017hockey}
Mehrnaz Fani, Helmut Neher, David~A Clausi, Alexander Wong, and John Zelek.
\newblock Hockey action recognition via integrated stacked hourglass network.
\newblock In {\em Proceedings of the IEEE Conference on Computer Vision and
  Pattern Recognition Workshops}, pages 29--37, 2017.

\bibitem{fragkiadaki2015recurrent}
Katerina Fragkiadaki, Sergey Levine, Panna Felsen, and Jitendra Malik.
\newblock Recurrent network models for human dynamics.
\newblock In {\em Proceedings of the IEEE International Conference on Computer
  Vision}, pages 4346--4354, 2015.

\bibitem{he2016deep}
Kaiming He, Xiangyu Zhang, Shaoqing Ren, and Jian Sun.
\newblock Deep residual learning for image recognition.
\newblock In {\em Proceedings of the IEEE conference on computer vision and
  pattern recognition}, pages 770--778, 2016.

\bibitem{hochreiter1997long}
Sepp Hochreiter and J{\"u}rgen Schmidhuber.
\newblock Long short-term memory.
\newblock {\em Neural computation}, 9(8):1735--1780, 1997.

\bibitem{hwang2017athlete}
Jihye Hwang, Sungheon Park, and Nojun Kwak.
\newblock Athlete pose estimation by a global-local network.
\newblock In {\em Proceedings of the IEEE Conference on Computer Vision and
  Pattern Recognition Workshops}, pages 58--65, 2017.

\bibitem{Iqbal_CVPR2017}
Umar Iqbal, Anton Milan, and Juergen Gall.
\newblock Posetrack: Joint multi-person pose estimation and tracking.
\newblock In {\em IEEE Conference on Computer Vision and Pattern Recognition
  (CVPR)}, 2017.

\bibitem{karpathy2014large}
Andrej Karpathy, George Toderici, Sanketh Shetty, Thomas Leung, Rahul
  Sukthankar, and Li Fei-Fei.
\newblock Large-scale video classification with convolutional neural networks.
\newblock In {\em Proceedings of the IEEE conference on Computer Vision and
  Pattern Recognition}, pages 1725--1732, 2014.

\bibitem{kay2017kinetics}
Will Kay, Joao Carreira, Karen Simonyan, Brian Zhang, Chloe Hillier, Sudheendra
  Vijayanarasimhan, Fabio Viola, Tim Green, Trevor Back, Paul Natsev, et~al.
\newblock The kinetics human action video dataset.
\newblock {\em arXiv preprint arXiv:1705.06950}, 2017.

\bibitem{kingma2014adam}
Diederik~P Kingma and Jimmy Ba.
\newblock Adam: A method for stochastic optimization.
\newblock {\em arXiv preprint arXiv:1412.6980}, 2014.

\bibitem{kuehne2011hmdb}
Hildegard Kuehne, Hueihan Jhuang, Est{\'\i}baliz Garrote, Tomaso Poggio, and
  Thomas Serre.
\newblock Hmdb: a large video database for human motion recognition.
\newblock In {\em 2011 International Conference on Computer Vision}, pages
  2556--2563. IEEE, 2011.

\bibitem{li2018jersey}
Gen Li, Shikun Xu, Xiang Liu, Lei Li, and Changhu Wang.
\newblock Jersey number recognition with semi-supervised spatial transformer
  network.
\newblock In {\em Proceedings of the IEEE Conference on Computer Vision and
  Pattern Recognition Workshops}, pages 1783--1790, 2018.

\bibitem{li2018resound}
Yingwei Li, Yi Li, and Nuno Vasconcelos.
\newblock Resound: Towards action recognition without representation bias.
\newblock In {\em Proceedings of the European Conference on Computer Vision
  (ECCV)}, pages 513--528, 2018.

\bibitem{liu2016spatio}
Jun Liu, Amir Shahroudy, Dong Xu, and Gang Wang.
\newblock Spatio-temporal lstm with trust gates for 3d human action
  recognition.
\newblock In {\em European Conference on Computer Vision}, pages 816--833.
  Springer, 2016.

\bibitem{reno2018convolutional}
Vito Reno, Nicola Mosca, Roberto Marani, Massimiliano Nitti, Tiziana D'Orazio,
  and Ettore Stella.
\newblock Convolutional neural networks based ball detection in tennis games.
\newblock In {\em Proceedings of the IEEE Conference on Computer Vision and
  Pattern Recognition Workshops}, pages 1758--1764, 2018.

\bibitem{russakovsky2015imagenet}
Olga Russakovsky, Jia Deng, Hao Su, Jonathan Krause, Sanjeev Satheesh, Sean Ma,
  Zhiheng Huang, Andrej Karpathy, Aditya Khosla, Michael Bernstein, et~al.
\newblock Imagenet large scale visual recognition challenge.
\newblock {\em International journal of computer vision}, 115(3):211--252,
  2015.

\bibitem{shukla2018automatic}
Pushkar Shukla, Hemant Sadana, Apaar Bansal, Deepak Verma, Carlos Elmadjian,
  Balasubramanian Raman, and Matthew Turk.
\newblock Automatic cricket highlight generation using event-driven and
  excitement-based features.
\newblock In {\em Proceedings of the IEEE Conference on Computer Vision and
  Pattern Recognition Workshops}, pages 1800--1808, 2018.

\bibitem{soomro2012ucf101}
Khurram Soomro, Amir~Roshan Zamir, and Mubarak Shah.
\newblock Ucf101: A dataset of 101 human actions classes from videos in the
  wild.
\newblock {\em arXiv preprint arXiv:1212.0402}, 2012.

\bibitem{sutskever2014sequence}
Ilya Sutskever, Oriol Vinyals, and Quoc~V Le.
\newblock Sequence to sequence learning with neural networks.
\newblock In {\em Advances in neural information processing systems}, pages
  3104--3112, 2014.

\bibitem{tran2015learning}
Du Tran, Lubomir Bourdev, Rob Fergus, Lorenzo Torresani, and Manohar Paluri.
\newblock Learning spatiotemporal features with 3d convolutional networks.
\newblock In {\em Proceedings of the IEEE international conference on computer
  vision}, pages 4489--4497, 2015.

\bibitem{tran2018closer}
Du Tran, Heng Wang, Lorenzo Torresani, Jamie Ray, Yann LeCun, and Manohar
  Paluri.
\newblock A closer look at spatiotemporal convolutions for action recognition.
\newblock In {\em Proceedings of the IEEE conference on Computer Vision and
  Pattern Recognition}, pages 6450--6459, 2018.

\bibitem{ul2018estimating}
Noor Ul~Huda, Kasper~H Jensen, Rikke Gade, and Thomas~B Moeslund.
\newblock Estimating the number of soccer players using simulation-based
  occlusion handling.
\newblock In {\em Proceedings of the IEEE Conference on Computer Vision and
  Pattern Recognition Workshops}, pages 1824--1833, 2018.

\bibitem{ullah2018directed}
Mohib Ullah and Faouzi Alaya~Cheikh.
\newblock A directed sparse graphical model for multi-target tracking.
\newblock In {\em Proceedings of the IEEE Conference on Computer Vision and
  Pattern Recognition Workshops}, pages 1816--1823, 2018.

\bibitem{veeriah2015differential}
Vivek Veeriah, Naifan Zhuang, and Guo-Jun Qi.
\newblock Differential recurrent neural networks for action recognition.
\newblock In {\em Proceedings of the IEEE international conference on computer
  vision}, pages 4041--4049, 2015.

\bibitem{wang2016temporal}
Limin Wang, Yuanjun Xiong, Zhe Wang, Yu Qiao, Dahua Lin, Xiaoou Tang, and Luc
  Van~Gool.
\newblock Temporal segment networks: Towards good practices for deep action
  recognition.
\newblock In {\em European conference on computer vision}, pages 20--36.
  Springer, 2016.

\bibitem{zhou2018temporal}
Bolei Zhou, Alex Andonian, Aude Oliva, and Antonio Torralba.
\newblock Temporal relational reasoning in videos.
\newblock In {\em Proceedings of the European Conference on Computer Vision
  (ECCV)}, pages 803--818, 2018.

\bibitem{zhou2015cnnlocalization}
B. Zhou, A. Khosla, Lapedriza. A., A. Oliva, and A. Torralba.
\newblock {Learning Deep Features for Discriminative Localization.}
\newblock {\em CVPR}, 2016.

\end{thebibliography}
\end{document}